\def\year{2021}\relax
\documentclass[letterpaper]{article} 
\usepackage{aaai21}  
\usepackage{times}  
\usepackage{helvet} 
\usepackage{courier}  
\usepackage[hyphens]{url}  
\usepackage{graphicx} 
\usepackage{natbib}  
\usepackage{caption} 
\usepackage[ampersand]{easylist}
\usepackage[utf8]{inputenc}
\usepackage[T1]{fontenc}
\usepackage[super]{nth}
\usepackage{listings}
\usepackage{dirtree}
\usepackage{amsmath}
\usepackage{subfig}
\usepackage{xcolor}

\lstdefinestyle{Python}{ 
    commentstyle=\color{orange},
    keywordstyle=\color{blue},
    numberstyle=\tiny\color{gray},
    stringstyle=\color{green},
    basicstyle=\scriptsize\ttfamily,
    breakatwhitespace=false,
    breaklines=true,
    captionpos=b,
    keepspaces=true,
    numbers=left,
    numbersep=5pt,
    showspaces=false,
    showstringspaces=false,
    showtabs=false,
    tabsize=2,
    morekeywords={self,np}
}
\newcommand{\ie}{\textit{i}.\textit{e}., }
\newcommand{\eg}{\textit{e}.\textit{g}. }

\title{Exploring reinforcement learning techniques for discrete\\and continuous control tasks in the MuJoCo environment}
\author{
    Vaddadi Sai Rahul \textsuperscript{\rm 1},
    Debajyoti Chakraborty \textsuperscript{\rm 1}
    \\
}
\affiliations{
    \textsuperscript{\rm 1}Northeastern University\\

    360 Huntington Avenue\\
    Boston, Massachusetts 02115\\
    vaddadi.s@northeastern.edu,
    chakraborty.de@northeastern.edu

}

\begin{document}

\maketitle

\begin{abstract}
We leverage the fast physics simulator, MuJoCo to run tasks in a continuous control environment and reveal details like the observation space, action space, rewards, etc. for each task. We benchmark value-based methods for continuous control by comparing Q-learning and SARSA through a discretization approach, and using them as baselines, progressively moving into one of the state-of-the-art deep policy gradient method DDPG. Over a large number of episodes, Q-learning outscored SARSA, but DDPG outperformed both in a small number of episodes. Lastly, we also fine-tuned the model hyper-parameters expecting to squeeze more performance but using lesser time and resources. We anticipated that the DDPG's new design would vastly improve performance, yet after only a few episodes, we were able to achieve decent average rewards. We expect to improve the performance provided adequate time and computational resources. Furthermore, without any modification, the methods were adapted to the other MuJoCo contexts.
\end{abstract}

\section{Introduction}
We will look at several reinforcement learning strategies for solving problems in discrete and continuous observation spaces, as well as discrete and continuous action spaces, in this work. In a continuous environment, predicting behaviors over a continuous space has always been a challenging challenge for an agent. Distinguishing the observation and action spaces is an obvious approach to these problems. The loss of information that occurs with dividing a continuous region into 'K' buckets is a significant constraint. Increasing the number of buckets can assist, but for continuous areas, it becomes intractable when the action-value table grows exponentially huge. We investigate the various environments in OpenAI gym's MuJoCo in this paper. To cope with continuous methods via bucketing, we use model-free temporal difference learning approaches - Q-learning and SARSA as a baseline. To improve the findings, the Deep Deterministic Policy Gradient (DDPG) was used.

We consider a typical reinforcement learning setup in which an agent interacts with its environment. The environment we have taken into consideration here, is \textbf{MuJoCo} (stands for \textbf{Mu}lti-\textbf{Jo}int dynamics with \textbf{Co}ntact). It is a general-purpose physics engine designed to help with research and development in robotics, biomechanics, machine learning, and other fields that need quick and precise modeling of articulated structures interacting with their surroundings.

\subsection{Environment.}
MuJoCo is a C/C++ library with a C API, which operates on low-level data structures which are pre-allocated by the built-in XML parser and compiler. Interactive visualization with a native GUI, produced in OpenGL, is included in the package. It provides continuous control tasks, running in a fast physics simulator and a plethora of utility functions for computing physics-related numbers.
MuJoCo offers a range of continuous control task:

\subsection{Model elements.}
The elements of a MuJoCo model are as:
\begin{enumerate}
    \item \textit{Body:} Bodies are the components that make up kinematic trees, \ie a tree of rigid bodies, such as, the human body, only having a predefined mass and inertia. Bodies do not posses any geometric properties.
    \item \textit{Joint:} Joints are defined inside bodies. Joints help to create motion between the particular body and its parent, otherwise they would be stiff and immovable. MuJoCo joints have four primitive types: \textsl{slide}, \textsl{hinge}, \textsl{ball} and \textsl{free}.
    \item \textit{DOF:} Degrees of freedom (DOFs) refers to the limits to which physical movement of the rigid bodies are possible. They are closely related to joints, however, different joints can have multiple DOFs. DOFs can have properties like damping, maximum velocity, armature, inertia, friction and other relevant data from coordinate systems.
    \item \textit{Geom:} Geoms are mass-less geometric objects primarily used in collision detection. MuJoCo supports geom types as \textsl{plane}, \textsl{sphere}, \textsl{capsule}, \textsl{ellipsoid}, \textsl{box}, \textsl{cone}, and \textsl{mesh}.
    \item \textit{Site:} Sites are locations of interest that are defined in the bodies' local frames and hence move with them. They are utilized in the engine to route tendons and apply various sorts of forces, but they may also be used by the application to encode sensor positions and other information.
    \item \textit{Constraint:} Constraints are used to define a set of pre-formulated rules that specify how they will behave in the environment, like restraining ball or hinge joints.
    \item \textit{Tendon:} A tendon can be used to impose constraints as "...the shortest path that passes through a sequence of specified sites or wraps around specified geoms."
    \item \textit{Actuator:} Actuators receive control inputs from the environment that directly co-relate to the movement or kinematics of the model. They can transmit forces (\eg torque), on any of joints, sites or tendons.
\end{enumerate}

\section{Related work}
Researchers have recently achieved substantial success by integrating deep learning capabilities for learning feature representations with reinforcement learning. Some instances include teaching agents to play video games using raw pixel data and teaching them sophisticated manipulation skills. Other instances include designing generalized agents that can "reinforce" itself into any task, given enough time and resources.

Expected SARSA might be employed for TD-learning approaches. However, due to the high spatial complexity, Tabular representations proved inefficient.

Another solution to our problem might be deep Q-learning. The features are calculated using a neural network in a deep form of approximate Q-learning. Despite the fact that it operates with continuous data, it models the probability distribution of discrete actions, necessitating the binning of the action space.

For our scenario, Actor-Critic, a policy gradient approach, might have also been employed. It performs effectively in areas where continuous control is required. However, the intended Q-value and present Q-value are both created by the same network, which is a huge disadvantage.The calculated TD error becomes inconsistent as a result of inconsistent weight changes.

In recent times, the state of the art in reinforcement learning in continuous control tasks are achieved in some or the other variation of deterministic policy gradient methods, \eg, Deep Deterministic Policy Gradient, Advantage Actor Critic (A2C), Asynchronous Advantage Actor Critic (A3C), Twin delayed deep deterministic policy gradient (TD3) etc.,.

\section{Background}
The tasks in the environment can be primarily associated with Locomotion, although few overlap with basic and hierarchical task as well, \ie, Locomotion + Food collection.

\subsection{Ant.}
The task is to make a 3-dimensional four-legged robot walk.
\begin{figure}[h]
    \centering
    \textbf{Ant-v2}\par\medskip
    \includegraphics[width=\columnwidth]{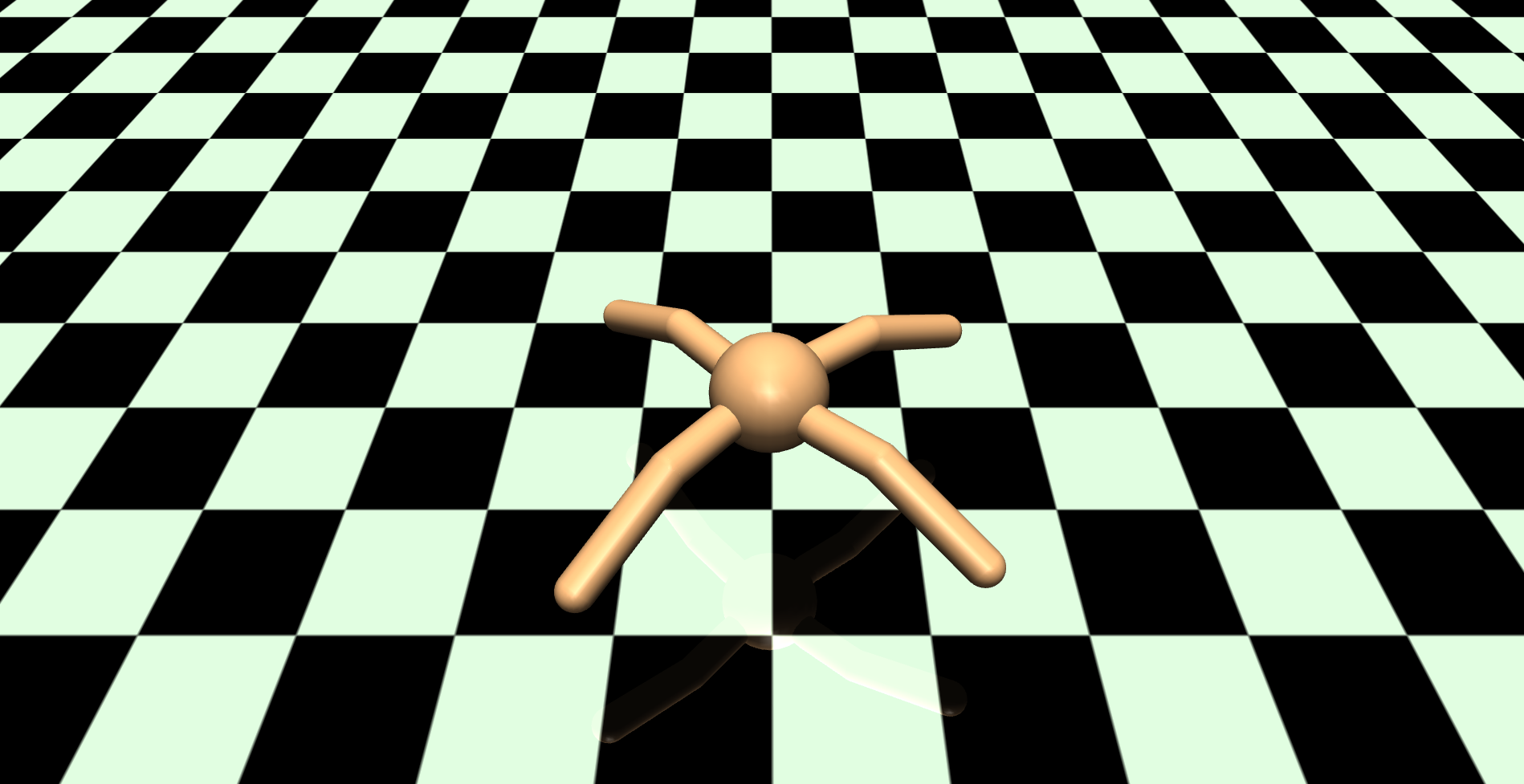}
    \caption{Four-legged ant navigating the environment}
\end{figure}
\subsubsection{XML description}
The ant has a spherical torso, with each of its four legs constituting of three capsule mesh geoms, connected by two hinge joints. The four legs are connected to the "torso" by four free joints.
\par\noindent\rule{\columnwidth}{0.1pt}
\dirtree{%
.1 World Body.
.2 Torso.
.3 Front left leg.
.4 Hip 1.
.5 Ankle 1.
.3 Front right leg.
.4 Hip 2.
.5 Ankle 2.
.3 Back leg.
.4 Hip 3.
.5 Ankle 3.
.3 Right back leg.
.4 Hip 4.
.5 Ankle 4.
}
\dirtree{%
.1 Actuator.
.2 Hip 1.
.2 Ankle 1.
.2 Hip 2.
.2 Ankle 2.
.2 Hip 3.
.2 Ankle 3.
.2 Hip 4.
.2 Ankle 4.
}
\par\noindent\rule{\columnwidth}{0.1pt}
\subsubsection{State space.} Has a shape of (111, ).
\begin{lstlisting}[frame=single]
 def _get_obs(self):
    return np.concatenate([
        self.sim.data.qpos.flat[2:],
        self.sim.data.qvel.flat,
        np.clip(self.sim.data.cfrc_ext, -1, 1).flat,
    ])
\end{lstlisting}
\begin{itemize}
    \item \textbf{self.sim.data.qpos} are the positions, with the first 7 element being the 3D position (x,y,z) and orientation (quaternion x,y,z,w) of the torso, and the remaining 8 positions being the joint angles.
    \item The \textbf{[2:], operation} removes the first 2 elements from the position \ie the X and Y position of the agent's torso.
    \item \textbf{self.sim.data.qvel} are the velocities, with the first 6 elements being the 3D velocity (x,y,z) and 3D angular velocity (x,y,z) and the remaining 8 are the joint velocities.
    \item The \textbf{cfrc\_ext} are the external forces (force x,y,z and torque x,y,z) applied to each of the links at the center of mass. This is 14 * 6: the ground link, the torso link plus the 12 links for all legs (3 links for each leg).
\end{itemize}
\subsubsection{Action space.} Has a shape of (8, ), translating directly as torque upon the 8 hinge joint actuators (2 for each leg).
\subsubsection{Rewards.} The rewards are represented as:
\begin{lstlisting}[frame=single]
ctrl_cost = self.control_cost(action)
contact_cost = self.contact_cost

forward_reward = x_velocity
healthy_reward = self.healthy_reward

rewards = forward_reward + healthy_reward
costs = ctrl_cost + contact_cost

reward = rewards - costs
\end{lstlisting}
\begin{itemize}
    \item Episodic reward is calculated by inflicting a cost on the total reward for the ant.
    \item One of the cost is a control cost for taking actions in the environment. Another is directly proportional to how many contacts the ant makes with the ground.
    \item This cost is deducted from the summed reward for moving forward and for being upright most of the time.
\end{itemize}

\subsection{HalfCheetah.}
The task is to make a 2-dimensional cheetah robot run.
\begin{figure}[h]
    \centering
    \textbf{HalfCheetah-v2}\par\medskip
    \includegraphics[width=\columnwidth]{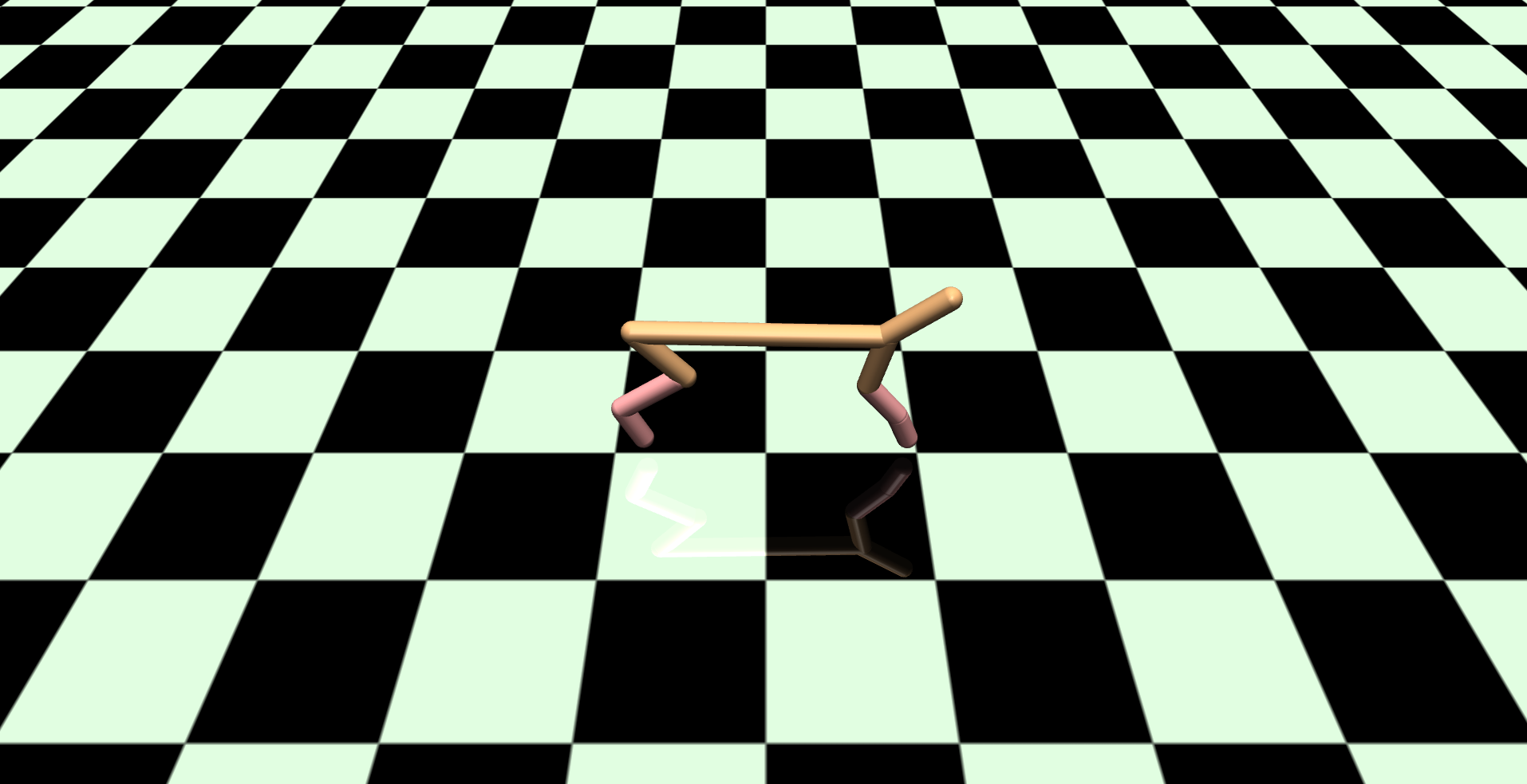}
    \caption{Two-legged cheetah running in the environment}
\end{figure}
\subsubsection{XML description}
The head and torso of the HalfCheetah are both capsule mesh geoms. Each thigh, shin and feet are capsules with a hinge joint connecting them together. This is obviously true for both the front and the back legs.
\par\noindent\rule{\columnwidth}{0.1pt}
\dirtree{%
.1 World Body.
.2 Head.
.2 Torso.
.3 Back thigh.
.4 Back shin.
.5 Back foot.
.3 Front thigh.
.4 Front shin.
.5 Front foot.
}
\dirtree{%
.1 Actuator.
.2 Back thigh.
.2 Back shin.
.2 Back foot.
.2 Front thigh.
.2 Front shin.
.2 Front foot.
}
\par\noindent\rule{\columnwidth}{0.1pt}
\subsubsection{State space} Has a shape of (17, ), as position and velocity for the slider joints, and angle and angular velocities for the hinge joints (3 for each leg, 3 axes for body).
\begin{center}
    \begin{tabular}{p{1.5cm}p{1.5cm}p{3.3cm}}
        \hline
            \textbf{Name}   &   \textbf{Joint}  &   \textbf{Parameter}  \\
        \hline
            rootx   &   slider  &   position (m)                \\
            rootz   &   slider  &   position (m)                \\
            rooty   &   hinge   &   angle (rad)                 \\
            bthigh  &   hinge   &   angle (rad)                 \\
            bshin   &   hinge   &   angle (rad)                 \\
            bfoot   &   hinge   &   angle (rad)                 \\
            fthigh  &   hinge   &   angle (rad)                 \\
            fshin   &   hinge   &   angle (rad)                 \\
            ffoot   &   hinge   &   angle (rad)                 \\
            rootx   &   slider  &   velocity (m/s)              \\
            rootz   &   slider  &   velocity (m/s)              \\
            rooty   &   hinge   &   angular velocity (rad/s)    \\
            bthigh  &   hinge   &   angular velocity (rad/s)    \\
            bshin   &   hinge   &   angular velocity (rad/s)    \\
            bfoot   &   hinge   &   angular velocity (rad/s)    \\
            fthigh  &   hinge   &   angular velocity (rad/s)    \\
            fshin   &   hinge   &   angular velocity (rad/s)    \\
            ffoot   &   hinge   &   angular velocity (rad/s)    \\
        \hline
    \end{tabular}
\end{center}
\subsubsection{Action space} Has a shape of (6, ), translating directly as torque upon the 2 hinge joint actuators (3 for each leg).
\begin{center}
    \begin{tabular}{p{1.5cm}p{1.5cm}p{2cm}}                        \\
        \hline
            \textbf{Name}   &   \textbf{Actuator}  &   \textbf{Parameter}  \\
        \hline
            bthigh  &   hinge   &   torque (Nm) \\
            bshin   &   hinge   &   torque (Nm) \\
            bfoot   &   hinge   &   torque (Nm) \\
            fthigh  &   hinge   &   torque (Nm) \\
            fshin   &   hinge   &   torque (Nm) \\
            ffoot   &   hinge   &   torque (Nm) \\
        \hline
    \end{tabular}
\end{center}
\subsubsection{Rewards.} The rewards are represented as:
\begin{lstlisting}[frame=single]
def control_cost(self, action):
    control_cost = self._ctrl_cost_weight * np.sum(np.square(action))
    return control_cost

ctrl_cost = self.control_cost(action)
forward_reward = self._forward_reward_weight * x_velocity
reward = forward_reward - ctrl_cost
\end{lstlisting}

\subsection{Humanoid.}
The task is to make a 3-dimensional two-legged robot walk.
\begin{figure}[h]
    \centering
    \textbf{Humanoid-v2}\par\medskip
    \includegraphics[width=\columnwidth]{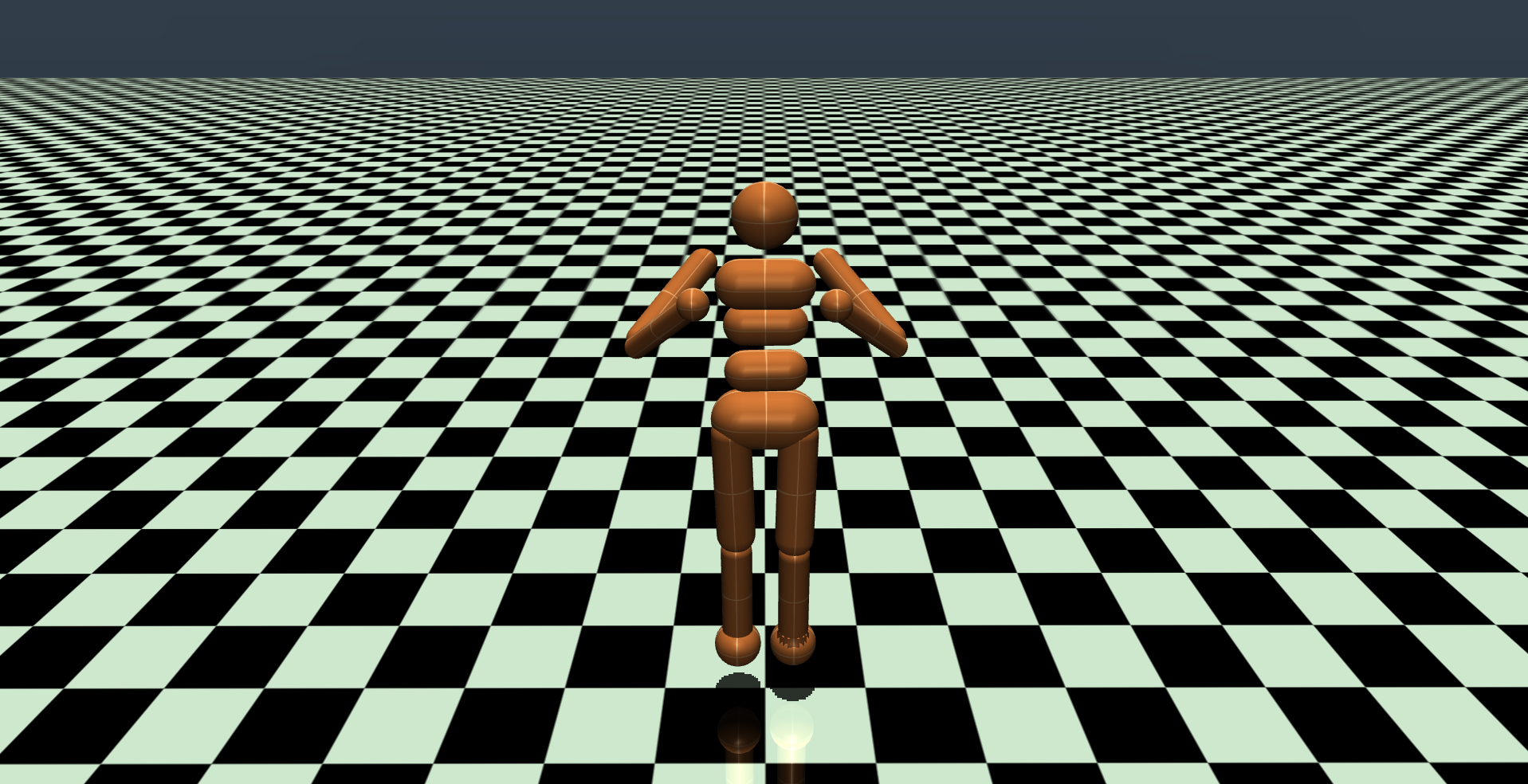}
    \caption{Two-legged humanoid learning how to walk}
\end{figure}
\subsubsection{XML description}
The head, torso and uwaist are sphere and two capsule geom meshes respectively. Also, conjoined are the left arm, right arm and the lower waist. Pelvis is a part of the lower waist which have the legs connected as a hinge joint. Both the 2 legs and 2 arms have 2 hinge joints each, responsible for moving both the lower arm and hand, and both the shin and foot respectively.
\par\noindent\rule{\columnwidth}{0.1pt}
\dirtree{%
.1 World Body.
.2 Head.
.2 Torso.
.2 Right upper arm.
.3 Right lower arm.
.4 Right hand.
.2 Left upper arm.
.3 Left lower arm.
.4 Left hand.
.2 Upper waist.
.2 Lower waist.
.3 Pelvis.
.4 Right thigh.
.5 Right shin.
.6 Right foot.
.4 Left thigh.
.5 Left shin.
.6 Left foot.
}
\dirtree{%
.1 Tendon.
.2 Left hip.
.2 Left knee.
.2 Right hip.
.2 Right knee.
}
\dirtree{%
.1 Actuator.
.2 Abdomen Y.
.2 Abdomen Z.
.2 Abdomen X.
.2 Right hip X.
.2 Right hip Z.
.2 Right hip Y.
.2 Right knee.
.2 Left hip X.
.2 Left hip Z.
.2 Left hip Y.
.2 Left knee.
.2 Right shoulder 1.
.2 Right shoulder 2.
.2 Right elbow.
.2 Left shoulder 1.
.2 Left shoulder 2.
.2 Left elbow.
}
\subsubsection{State space.} Has a shape of (376, ).
\begin{lstlisting}[frame=single]
def _get_obs(self):
    position = self.sim.data.qpos.flat.copy()
    velocity = self.sim.data.qvel.flat.copy()

    com_inertia = self.sim.data.cinert.flat.copy()
    com_velocity = self.sim.data.cvel.flat.copy()

    actuator_forces = self.sim.data.qfrc_actuator.flat.copy()
    external_contact_forces = self.sim.data.cfrc_ext.flat.copy()

    if self._exclude_current_positions _from_observation:
        position = position[2:]

    return np.concatenate(
        (
            position,
            velocity,
            com_inertia,
            com_velocity,
            actuator_forces,
            external_contact_forces,
        )
    )
\end{lstlisting}
\begin{itemize}
    \item \textbf{self.sim.data.qpos} are the positions, with the first 7 element being the 3D position (x,y,z) and orientation (quaternion x,y,z,w) of the torso, and the remaining 8 positions being the joint angles.
    \item The \textbf{[2:], operation} removes the first 2 elements from the position \ie the X and Y position of the agent's torso.
    \item \textbf{self.sim.data.qvel} are the velocities, with the first 6 elements being the 3D velocity (x,y,z) and 3D angular velocity (x,y,z) and the remaining 8 are the joint velocities.
    \item The \textbf{cfrc\_ext} are the external forces (force x,y,z and torque x,y,z) applied to each of the links at the center of mass. This is 14 * 6: the ground link, the torso link plus the 12 links for all legs (3 links for each leg).
    \item \textbf{qfrc\_actuator} are likely the actuator forces. \textbf{cinert} seems the center of mass based inertia and \textbf{cvel} the center of mass based velocity.
\end{itemize}
\subsubsection{Action space.} Has a shape of (17, ), translating directly as torque upon the 17 hinge joint actuators listed in the tree.

\subsubsection{Reward.} Represented same as for the Ant agent.

\subsection{InvertedDoublePendulum}
The task is to balance a pole on a pole, on a cart.
\begin{figure}[h]
    \centering
    \textbf{InvertedDoublePendulum-v2}\par\medskip
    \includegraphics[width=\columnwidth]{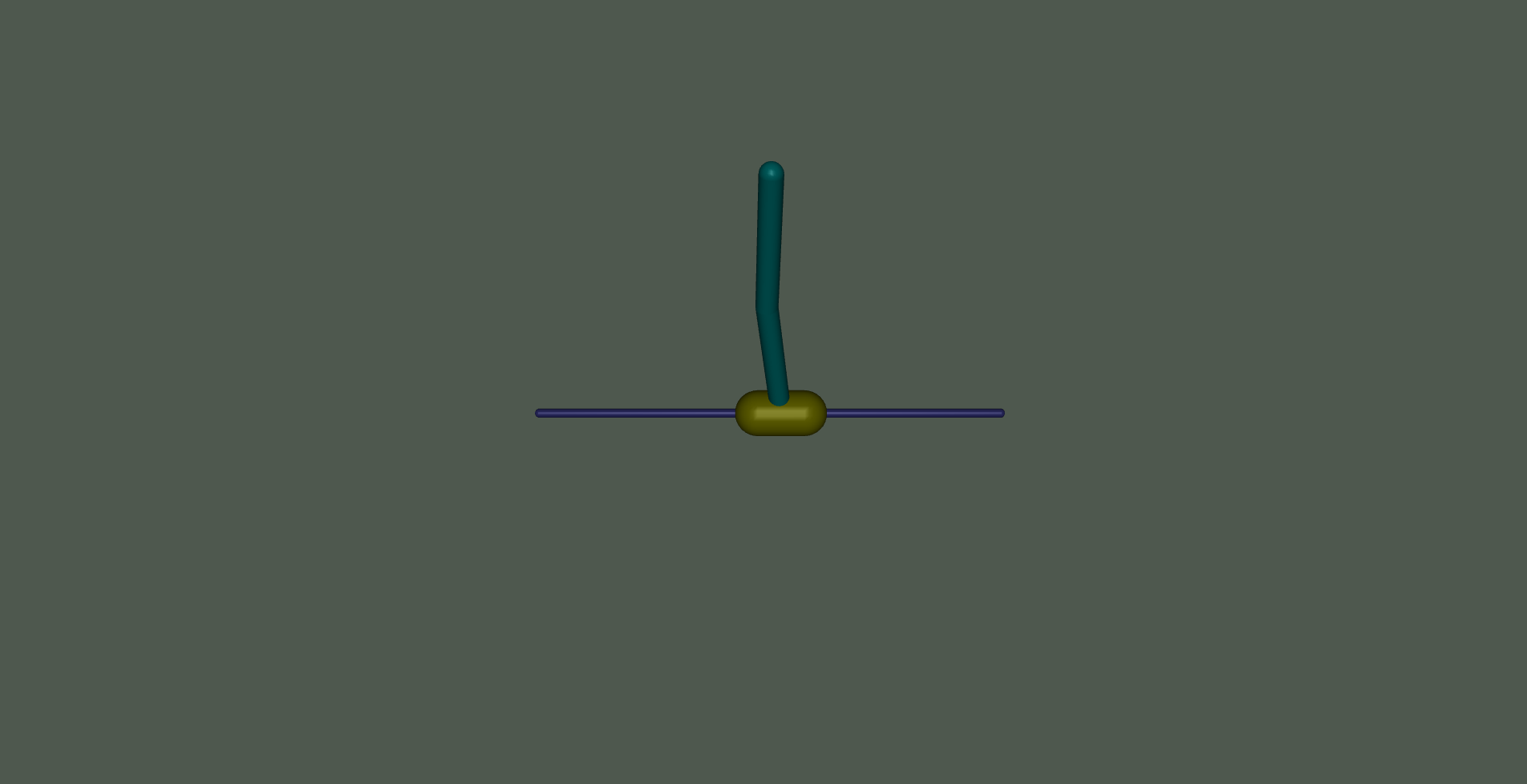}
    \caption{Cart balancing the pole in the environment}
\end{figure}
\subsubsection{XML description}
The root of this model is a capsule geom on a rail of joint type slide. It is connected to a pole of geom type capsule which in turn, is connected to another pole of geom type capsule via a hinge joint.
\par\noindent\rule{\columnwidth}{0.1pt}
\dirtree{%
.1 World body.
.2 Cart.
.3 Pole.
.4 Tip.
}
\dirtree{%
.1 Actuator.
.2 Slide.
}
\subsubsection{State space.} Has a shape of (11, ), represented as slider position and velocity for the cart, and angle and angular velocity for the two pole joints.
\begin{lstlisting}[frame=single]
def _get_obs(self):
    return np.concatenate(
        [
            self.sim.data.qpos[:1],
            np.sin(self.sim.data.qpos[1:]),
            np.cos(self.sim.data.qpos[1:]),
            np.clip(self.sim.data.qvel, -10, 10),
            np.clip(self.sim.data.qfrc_constraint, -10, 10),
        ]
    ).ravel()
\end{lstlisting}
\begin{center}
    \begin{tabular}{p{1.5cm}p{1.5cm}p{3.3cm}}
        \hline
            \textbf{Name}   &   \textbf{Joint}  &   \textbf{Parameter}  \\
        \hline
            cart    &   slider  &   position (m)                \\
            pole    &   hinge   &   angle (rad)                 \\
            cart    &   slider  &   velocity (m)                \\
            pole    &   hinge   &   angular velocity (rad/s)    \\
        \hline
    \end{tabular}
\end{center}
\subsubsection{Action space.} Has a shape of (1, ), represented as force in X-axis on the cart, resulting it to translate on the rail.
\begin{center}
    \begin{tabular}{p{1.5cm}p{1.5cm}p{2cm}}                        \\
        \hline
            \textbf{Name}   &   \textbf{Actuator}  &   \textbf{Parameter}  \\
        \hline
            cart    &   motor   &   force x (N)     \\
        \hline
    \end{tabular}
\end{center}
\subsubsection{Reward.} The rewards are represented as:
\begin{lstlisting}[frame=single]
dist_penalty = 0.01 * x ** 2 + (y - 2) ** 2
v1, v2 = self.sim.data.qvel[1:3]
vel_penalty = 1e-3 * v1 ** 2 + 5e-3 * v2 ** 2
alive_bonus = 10
r = alive_bonus - dist_penalty - vel_penalty
\end{lstlisting}

\subsection{Reacher.}
The task is to make a 2-dimensional robot reach to a randomly located target.
\begin{figure}[h]
    \centering
    \textbf{Reacher-v2}\par\medskip
    \includegraphics[width=\columnwidth]{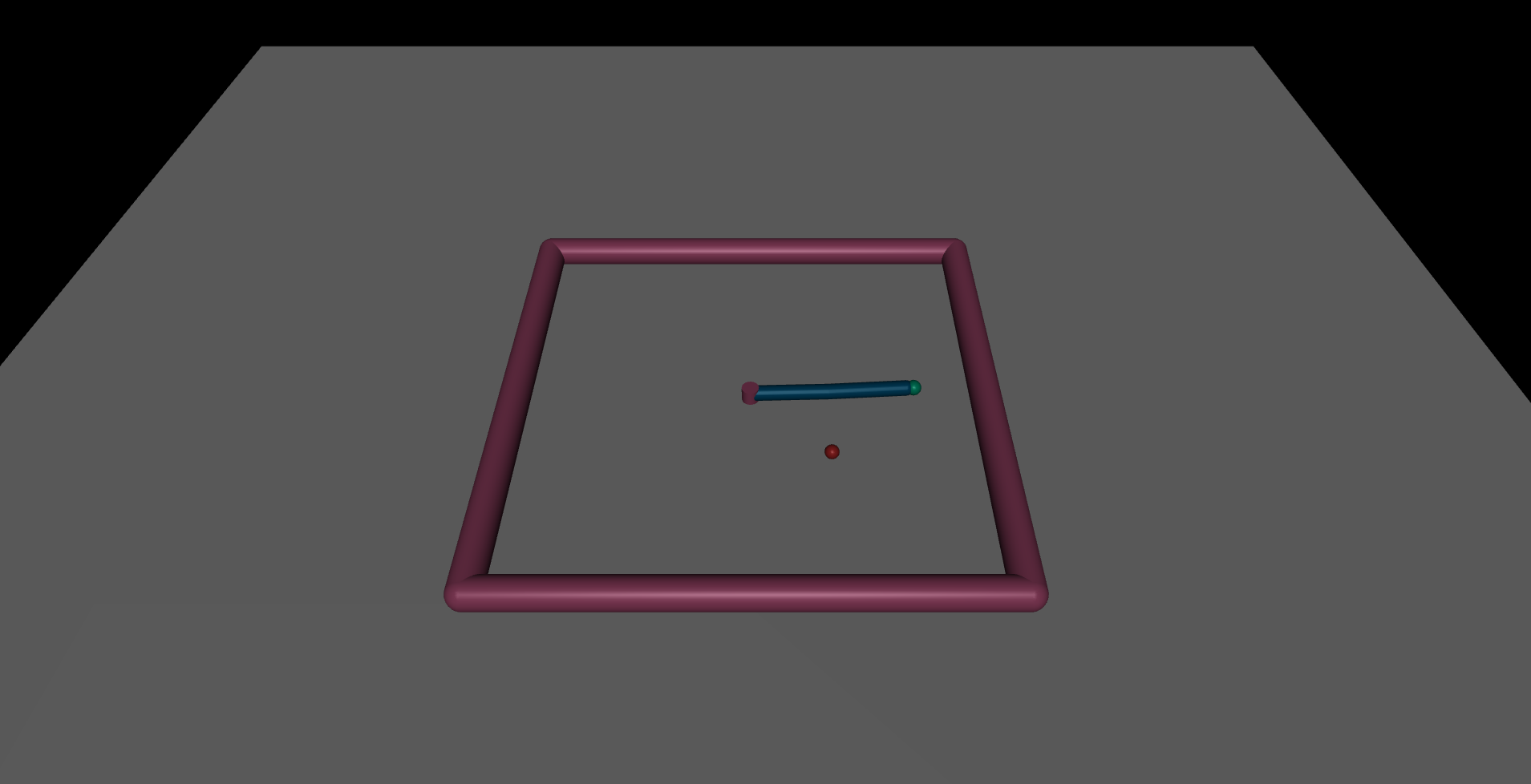}
    \caption{Robot trying to reach the randomly generated target in the environment}
\end{figure}
\subsubsection{XML description}
The "arm" of the agent consists of two rigid capsule geoms connected via hinge joint and ending in a sphere geom, termed the "fingertip". The environment also contains the target as a spherical geom which the agent expects to reach to and also, four capsule geom "arena" which does not directly affect the agent's performance.
\par\noindent\rule{\columnwidth}{0.1pt}
\dirtree{%
.1 World body.
.2 Body 0.
.3 Body 1.
.4 Fingertip.
.2 Target.
}
\dirtree{%
.1 Actuator.
.2 Joint 0.
.2 Joint 1.
}
\subsubsection{State space.} Has a shape of (11, ).
\begin{lstlisting}[frame=single]
def get_body_com(self, body_name):
    return self.data.get_body_xpos(body_name)

def _get_obs(self):
    theta = self.sim.data.qpos.flat[:2]
    return np.concatenate(
        [
            np.cos(theta),
            np.sin(theta),
            self.sim.data.qpos.flat[2:],
            self.sim.data.qvel.flat[:2],
            self.get_body_com("fingertip") - self.get_body_com("target"),
        ]
    )
\end{lstlisting}
\begin{center}
    \begin{tabular}{p{1.5cm}p{1.5cm}p{3.3cm}}
        \hline
            \textbf{Name}   &   \textbf{Joint}  &   \textbf{Parameter}  \\
        \hline
            joint0      &   hinge   &   angle (rad)                 \\
            joint1      &   hinge   &   angle (rad)                 \\
            joint0      &   hinge   &   angular velocity (rad/s)    \\
            joint1      &   hinge   &   angular velocity (rad/s)    \\
            target      &   slider  &   position (m)                \\
        \hline
    \end{tabular}
\end{center}
\subsubsection{Action space.} Has a shape of (2, ), represented as torque on the two joints, resulting in the agent reaching the target.
\begin{center}
    \begin{tabular}{p{1.5cm}p{1.5cm}p{2cm}}                        \\
        \hline
            \textbf{Name}   &   \textbf{Actuator}  &   \textbf{Parameter}  \\
        \hline
            joint0  &   motor   &   torque (Nm)     \\
            joint1  &   motor   &   torque (Nm)     \\
        \hline
    \end{tabular}
\end{center}
\subsubsection{Reward.} The rewards are represented as:
\begin{lstlisting}[frame=single]
vec = self.get_body_com("fingertip") - self.get_body_com("target")
reward_dist = -np.linalg.norm(vec)
reward_ctrl = -np.square(a).sum()
reward = reward_dist + reward_ctrl
\end{lstlisting}

\subsection{Swimmer.}
The task is to make a 2-dimensional robot swim.
\begin{figure}[h]
    \centering
    \textbf{Swimmer-v2}\par\medskip
    \includegraphics[width=\columnwidth]{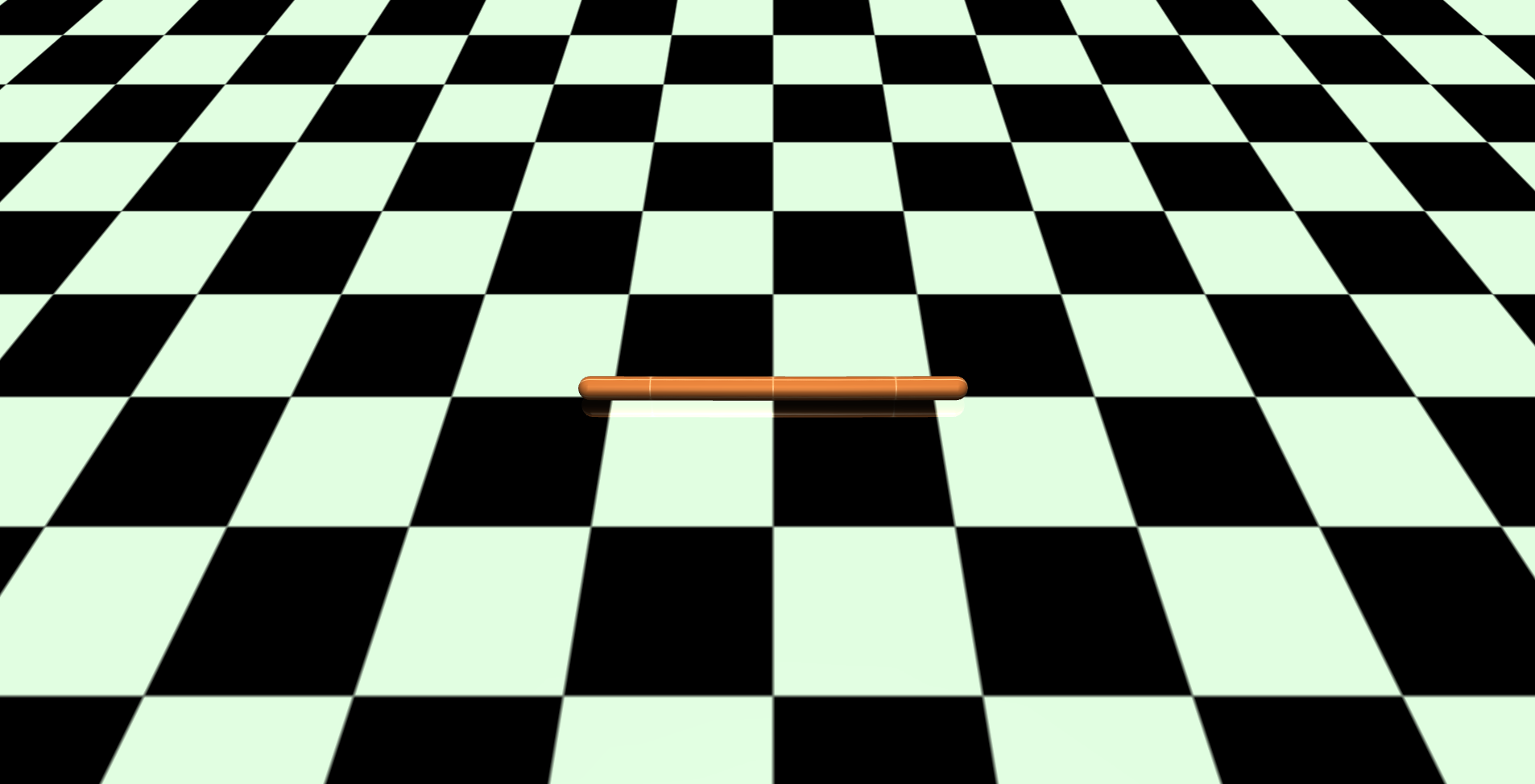}
    \caption{Robot trying to swim in the environment}
\end{figure}
\subsubsection{XML description.}
The torso of the agent has two rigid capsule geoms, connected over two hinge joints. 
\par\noindent\rule{\columnwidth}{0.1pt}
\dirtree{%
.1 World body.
.2 Torso.
.3 Mid.
.4 Back.
}
\dirtree{%
.1 Actuator.
.2 Rot 2.
.2 Rot 3.
}
\subsubsection{State space.} Has a shape of (8, ).
\begin{lstlisting}[frame=single]
def _get_obs(self):
    position = self.sim.data.qpos.flat.copy()
    velocity = self.sim.data.qvel.flat.copy()

    if self._exclude_current_positions_from_observation:
        position = position[2:]

    observation = np.concatenate([position, velocity]).ravel()
    return observation
\end{lstlisting}
\begin{center}
    \begin{tabular}{p{1.5cm}p{1.5cm}p{3.3cm}}
        \hline
            \textbf{Name}   &   \textbf{Joint}  &   \textbf{Parameter}  \\
        \hline
            slider1     &   slide   &   position (m)                \\
            slider2     &   slide   &   position (m)                \\
            slider1     &   slide   &   velocity (m/s)              \\
            slider2     &   slide   &   veloctiy (m/s)              \\
            rot2        &   hinge   &   angle (rad)                 \\
            rot3        &   hinge   &   angle (rad)                 \\
            rot2        &   hinge   &   angular velocity (rad/s)    \\
            rot3        &   hinge   &   angular velocity (rad/s)    \\
        \hline
    \end{tabular}
\end{center}
\subsubsection{Action space.} Has a shape of (2, ), represented as torque on the two joints, resulting in the agent "swimming".
\begin{center}
    \begin{tabular}{p{1.5cm}p{1.5cm}p{2cm}}                        \\
        \hline
            \textbf{Name}   &   \textbf{Actuator}  &   \textbf{Parameter}  \\
        \hline
            rot2    &   motor   &   torque (Nm) \\
            rot3    &   motor   &   torque (Nm) \\
        \hline
    \end{tabular}
\end{center}
\subsubsection{Reward.} The rewards are represented as:
\begin{lstlisting}[frame=single]
def control_cost(self, action):
    control_cost = self._ctrl_cost_weight * np.sum(np.square(action))
    return control_cost

xy_position_before = self.sim.data.qpos[0:2].copy()
xy_position_after = self.sim.data.qpos[0:2].copy()

xy_velocity = (xy_position_after - xy_position_before) / self.dt
x_velocity, y_velocity = xy_velocity

forward_reward = self._forward_reward_weight * x_velocity

ctrl_cost = self.control_cost(action)
reward = forward_reward - ctrl_cost
\end{lstlisting}

\subsection{Hopper.}
The task is to make a 2-dimensional robot hop.
\begin{figure}[h]
    \centering
    \textbf{Hopper-v2}\par\medskip
    \includegraphics[width=\columnwidth]{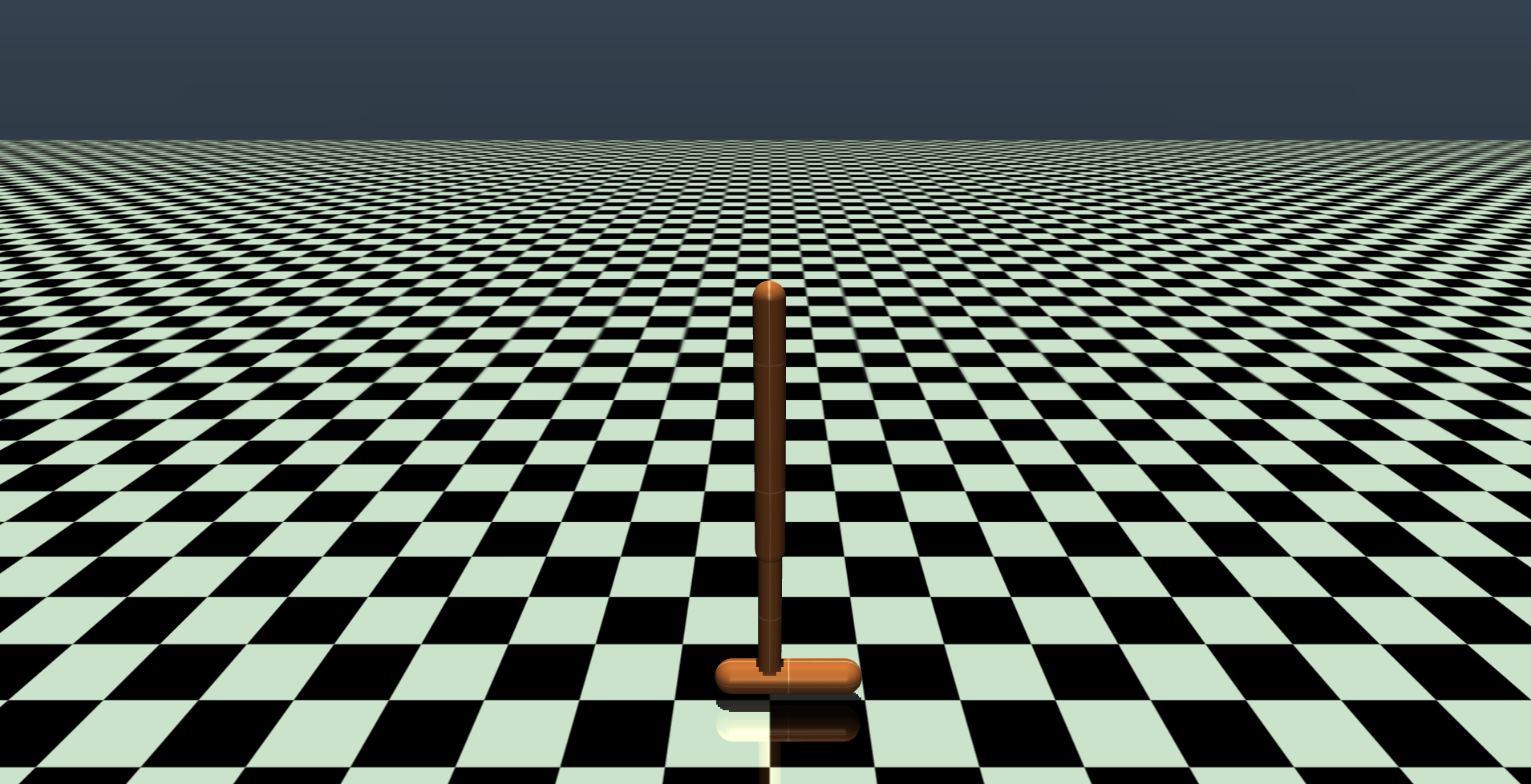}
    \caption{One-legged robot learning to hop}
\end{figure}
\subsubsection{XML description.}
The torso of the agent sequences a single thigh, a single leg and a single foot, all of them being mesh capsule geoms, and all connected via a hinge joint.
\par\noindent\rule{\columnwidth}{0.1pt}
\dirtree{%
.1 World Body.
.2 Torso.
.3 Thigh.
.4 Leg.
.5 Foot.
}
\dirtree{%
.1 Actuator.
.2 Thigh joint.
.2 Leg joint.
.2 Foot joint.
}
\subsubsection{State space.} Has a shape of (11, ).
\begin{lstlisting}[frame=single]
def _get_obs(self):
    position = self.sim.data.qpos.flat.copy()
    velocity = np.clip(self.sim.data.qvel.flat.copy(), -10, 10)

    if self._exclude_current_positions_from_observation:
        position = position[1:]

    observation = np.concatenate((position, velocity)).ravel()
    return observation
\end{lstlisting}
\begin{center}
    \begin{tabular}{p{1.5cm}p{1.5cm}p{3.3cm}}
        \hline
            \textbf{Name}   &   \textbf{Joint}  &   \textbf{Parameter}  \\
        \hline
            rootx   &   slider  &   position (m)                \\
            rootz   &   slider  &   position (m)                \\
            rooty   &   hinge   &   angle (rad)                 \\
            thigh   &   hinge   &   angle (rad)                 \\
            leg     &   hinge   &   angle (rad)                 \\
            foot    &   hinge   &   angle (rad)                 \\
            rootx   &   slider  &   velocity (m/s)              \\
            rootz   &   slider  &   velocity (m/s)              \\
            rooty   &   hinge   &   angular velocity (rad/s)    \\
            thigh   &   hinge   &   angular velocity (rad/s)    \\
            leg     &   hinge   &   angular velocity (rad/s)    \\
            foot    &   hinge   &   angular velocity (rad/s)    \\
        \hline
    \end{tabular}
\end{center}
\subsubsection{Action space.} Has a shape of (3, ), represented as torque on the three joints, resulting in the agent reaching the target.
\begin{center}
    \begin{tabular}{p{1.5cm}p{1.5cm}p{2cm}}                        \\
        \hline
            \textbf{Name}   &   \textbf{Actuator}  &   \textbf{Parameter}  \\
        \hline
            thigh\_joint     &   motor   &   torque (Nm)     \\
            leg\_joint       &   motor   &   torque (Nm)     \\
            foot\_joint      &   motor   &   torque (Nm)     \\
        \hline
    \end{tabular}
\end{center}
\subsubsection{Reward.} The rewards are represented as:
\begin{lstlisting}[frame=single]
x_position_before = self.sim.data.qpos[0]
x_position_after = self.sim.data.qpos[0]
x_velocity = (x_position_after - x_position_before) / self.dt

ctrl_cost = self.control_cost(action)

forward_reward = self._forward_reward_weight * x_velocity
healthy_reward = self.healthy_reward

rewards = forward_reward + healthy_reward
reward = rewards - ctrl_cost
\end{lstlisting}

\section{Project description}
\subsection{Online Value-Based Methods}
“Bootstrapping” in reinforcement learning means that the estimate of one state $V_\pi(s)$ builds upon the estimate of successor states $V_\pi(s')$. Dynamic programming uses bootstrapping and is a model-based learning. Other methods do not rely on bootstrapping and are known as model-free methods like Monte-Carlo. Temporal difference learning combines Monte-Carlo (model-free) and Dynamic programming (model-based).

The Temporal difference (TD) error is given by:
\begin{equation*}
    \delta_t = R_{t+1} + \gamma * V(s_{t+1}) - V(s_t)
\end{equation*}

$\delta_t$: TD error \\
$V(s_t)$: value estimate of state ‘$s_t$’ \\
$V(s_{t+1})$: value estimate of next state ‘$s_{t+1}$’ \\
$R_{t+1}$: reward obtained on transition from ‘$s_t$’ to ‘$s_{t+1}$’

\subsubsection{SARSA.} SARSA combines Generalized Policy Iteration with Temporal Difference learning to find improved policies. It uses action-values (Q-value) form of TD. The name ‘SARSA’ stands for $S_t,A_t,R_{t+1},S_{t+1},A_{t+1} \rightarrow (state, action, reward, next state, next action)$. It is an on-policy TD control method. \\
The update equation used by SARSA is:
\begin{equation*}
    Q(s_t,a_t) \Leftarrow Q(s_t,a_t) + \alpha * [R_{t+1} + \gamma * Q(s_{t+1},a_{t+1}) - Q(s_t,a_t)]
\end{equation*}

$Q(s_t,a_t)$: action-value estimate for state $s_t$ and action $a_t$
$\alpha$: learning rate
$\gamma$: discount factor
$Q(s_{t+1},a_{t+1})$: action-value estimate for state $s_{t+1}$ and action $a_{t+1}$

In Generalized policy iteration with SARSA, we continually estimate $Q_\pi$ for the behavior policy $\pi$, at the same time change $\pi$ towards greediness with respect to $Q_\pi$.

\subsubsection{Q-Learning.}
Q-learning is an off-policy Temporal difference control algorithm. In Q-learning, the incremental update is given by
\begin{align*}
    & Q(s_t,a_t) \Leftarrow Q(s_t,a_t) + \\
    & alpha * [R_{t+1} + \gamma * \max_aQ(s_{t+1},a) - Q(s_t,a_t)]
\end{align*}
$\alpha$: learning rate \\
$\gamma$: discount factor \\
$Q(s_t,a_t)$: action-value estimate for state $s_t$ and action $a_t$ \\
$Q(s_{t+1},a_{t+1})$: action-value estimate for state $s_{t+1}$ and action $a_{t+1}$ \\

The target policy is $\pi^* = {\rm argmax}_aQ(s,a)$. The term $\max_aQ(s_{t+1},a)$ selects greedy actions irrespective of actual policy $\pi$ (behavior policy).

\subsection{Policy gradient}
Until now, we considered action-value estimates for learning an optimal policy. As the observation and action spaces tend to grow, tabular methods prove inefficient due to the exponential growth of Q-table size, resulting in the curse of dimensionality problem. Now, we consider the class of methods that can select actions without using a value function. These are called as policy gradient methods. This method is applicable for learning optimal policies in the continuous observation space, using probability distributions over the action space.

For this, we use a parameterized policy given by  $\pi(a|s,\theta)=Pr\{A_t=a|S_t=s,\theta_t=\theta\}$ where, $'\theta'$ is the policy’s parameter vector. Like the weight parameter vector ‘w’ we use for approximate action-value functions $\hat{q}(s,a,w)$, here we use '$\theta$. The constraints on policy parameterization are:
\begin{equation*}
    \pi(a|s,\theta) \geq 0 \; \forall \; a \; \in \; A \; \textit{and} \; s \; \in \; S
\end{equation*}
\begin{equation*}
    \Sigma_{a \epsilon A} \pi (a|s,\theta) = 1 \; \forall \; s \; \in \; S
\end{equation*}

$\pi(a|s,\theta)$ is differentiable with respect to parameter $'\theta' \; \ie \; \nabla \; \pi(a|s,\theta)$ exists. In order to satisfy these conditions, we use a "softmax policy parameterization".

\begin{equation*}
    \pi(a|s,\theta) \; = \; e^{h(a,s,\theta)} \; / \; \Sigma_{b \in A} e^{h(b,s,\theta)}
\end{equation*}

$h(s,a,\theta)$ is knows as parameterized numerical preferences where $h(s,a,\theta) \in R$. The action with the highest preferences in each state are given the highest probabilities of being selected according to equation. Numerical preferences can be computed by a deep Artificial Neural Network, where $\theta$ is the vector of all connection weights of the network or could simply be linear in features.

$h(s,a,\theta) = \theta^T X(s,a)$ where X(s,a) is some feature vector.

The goal of RL is maximizing rewards in the long run $R_{t},R_{(t+1)},R_{(t+2)}...$. In the policy gradient case, our objective maximizing the average reward $r_\pi$ hence, we use gradient ascent.
\begin{equation*}
    r(\pi) = \Sigma_s \mu(s) \Sigma_a \pi(a|s,\theta) \Sigma_{s',r}p(s',r|s,a) * r
\end{equation*}
\begin{equation*}
        \nabla_\theta r(\pi) = \nabla_\theta [\Sigma_s \mu(s) \Sigma_a \pi(a|s,\theta) \Sigma_{s',r}p(s',r|s,a) * r]
\end{equation*}

From the product rule of calculus,

\begin{align*}
    \nabla_\theta r(\theta) &= \Sigma_s \mu(s) \nabla_\theta \Sigma_a \pi(a|s,\theta) \Sigma_{s',r}p(s',r|s,a) * r \\
    &+ \Sigma_s \nabla_\theta \mu(s) \Sigma_a \pi(a|s,\theta) \Sigma_{s',r}p(s',r|s,a) * r
\end{align*}

The challenge of this method lies in computing the gradient of the state distribution $\pi(s)$ as it changes with $\theta$. To address this, we use the “policy gradient theorem” which returns a simplified expression independent of $\nabla_\theta \mu(s)$.

\subsection{Deep Deterministic Policy Gradient}
Earlier method worked well with discrete action spaces but fails for continuous control problems. Deep Deterministic Policy Gradient (DDPG) incorporates Deterministic Policy Gradient (DPG) into the Actor-Critic structure to extend to continuous action spaces. It relies on off-policy updates using target networks.

DDPG makes use of 4 networks in total – \textbf{actor network}, \textbf{critic network}, \textbf{target-actor network}, and \textbf{target-critic network}. The actor network computes the deterministic policy $a_t = \mu(s_t|\theta^\mu)$, where $\theta^\mu$ are the weights for the actor network. However, this policy might not explore the full state and action space. To encourage exploration, it makes use of a random process called the Ornstein-Uhlenbeck Noise $N_t$. In a continuous setting, it is defined as:

\begin{equation*}
    dN_t = \beta * (\mu - N_t) * dt + \sigma * dW_t
\end{equation*}
In the discrete case,
\begin{equation*}
    N_{t+1} = (1 - \beta) * N_t - \mu + \sigma * (W_{t+1}  - W_t)
\end{equation*}

$N_t$ : noise at time ‘t’ \\
$\beta$ : decay or growth rate of the system \\
$\mu$ :  asymptotic mean \\
$\sigma$ : variation or size of noise \\
$W$ : wiener process \\

The Weiner process also known as Brownian motion is a stationary process with white noise increments of a noise distribution $N_t$ with $\mu = 0$ and $\sigma = 1$.

The Critic network $Q(s_t,a_t|\theta^Q)$ evaluates state-action pairs where $\theta^Q$ are its weights. The target actor and critic network denoted by $Q'$ and $\mu'$ with weights $\theta^{Q'}$ and $\theta^{\mu'}$ respectively are a soft copy of the weights of actor and critic network $\theta^\mu$ and $\theta^Q$ respectively.
\begin{equation*}
    \theta^{Q'} \Leftarrow \theta^Q
\end{equation*}
\begin{equation*}
    \theta^{\mu'} \Leftarrow \theta^\mu
\end{equation*}

Replay Buffer $R$ stores the transition dynamics of the environment \ie $R = \{(s_t,a_t,r_t,s_{t+1})\} \; \forall \; a_t \in A$ \textit{and} $s_t \in S; t \in [1,M']$ where $M'$ is the memory limit. Whenever an agent takes an action in the environment, the transition tuple $(s_t,a_t,r_t,s_{t+1})$ is added to the replay buffer. The objective of the critic network is minimizing the temporal difference between the target-critic network’s output and the estimated Q-value from its network.
\begin{equation*}
    y_i = r_i + \gamma * Q'(s_{i+1},\mu'(s_{i+1}|\theta^{\mu'})|\theta^{Q'})
\end{equation*}
\begin{equation*}
    L = (1/n) * (y_i - Q(s_i,\mu(s_i|\theta^\mu)|\theta^Q))
\end{equation*}

$\mu'(s_{i+1}|\theta^{\mu})$: estimated target-actor network’s policy \\
$\mu(s_i|\theta^\mu)|\theta^Q)$: estimated actor network’s policy \\
$Q'(s_{i+1},\mu'(s_{i+1}|\theta^{\mu'})|\theta^{Q'})$: estimated target-critic network’s Q-value \\
$Q(s_i,\mu(s_i|\theta^\mu)|\theta^Q)$: estimated critic network’s Q-value \\
$n$: number of random samples from replay buffer \\
$L$: critic loss \\
	
The objective of the actor network is to learn the optimal policy that maximizes the expected return. It uses the policy gradient to achieve its goal.
\begin{equation*}
    J(\theta) = E[Q(s,a)|s=s_t,a_t=\mu(s_t)]
\end{equation*}
\begin{equation*}
    \nabla_{\theta^\mu} J(\theta) \approx \nabla_a Q(s,a)\nabla_(\theta^\mu) \mu(s|\theta^\mu)
\end{equation*}

Across ‘n’ mini-batch samples from replay buffer,
\begin{align*}
    \nabla_{\theta^\mu} J(\theta) &\approx (1/n) \; * \\
    & \Sigma_i \nabla_a Q(s,a|\theta^Q)|_{s=s_i,a=\mu(s_i)} \nabla_{\theta^\mu} \mu(s|\theta^\mu)|_{s_i}
\end{align*}

The target networks are updated using a moving average equation with parameter '$\tau$', which indicates the fraction of weights carried over from the original actor-critic networks to the corresponding target networks.
\begin{equation*}
    \theta^{Q'} \Leftarrow \tau * \theta^Q + (1 - \tau) * \theta^{Q'}
\end{equation*}
\begin{equation*}
    \theta^{\mu'} \Leftarrow \tau * \theta^\mu + (1 - \tau) * \theta^{\mu'}
\end{equation*}

The original pseudo-code for DDPG is illustrated below:
\begin{figure}[h]
    \centering
    \textbf{Deep Deterministic Policy Gradient (DDPG)}\par\medskip
    \includegraphics[width=\columnwidth]{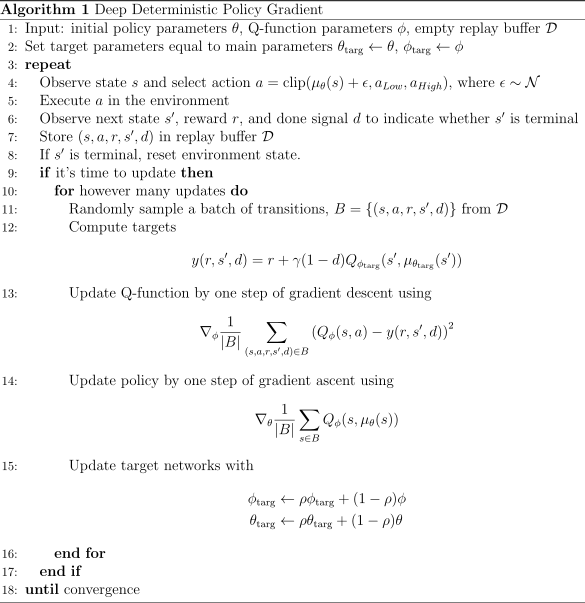}
    \caption{Pseudo-code from the original paper}
\end{figure}

\section{Experiments}
All the experiments were run on a Nvidia GeForce GTX 1060 with Max-Q Design and Intel Core i7-7700HQ CPU, with a physical memory (RAM) size of 16 GB. Tabular Q-learning and SARSA (State–action–reward–state–action) were the baseline methods chosen. Our initial approach was to experiment with the performance of discrete observation and action space methods on continuous observation and control tasks.  As the ranges were [-inf, inf] for each observation, we sampled across 10k observations and clipped the maximum and minimum ranges to [-25, 25]. We discretized the continuous values into 2 buckets categorized into {0, 1}, for both the action and observation spaces. We varied the learning rate starting from {0.2, 0.3, …, 0.9}. Other parameters chosen were: $\gamma = 0.99$, number of episodes (epochs) = 500 and number of steps per episode = 1000. Actions were selected using an epsilon-greedy policy with $\epsilon=0.99$ decaying at a rate of:
\begin{equation*}
\epsilon = log_{10} ((e^\epsilon + 1) / 25)
\end{equation*}
The following curves were observed for Tabular Q-learning and SARSA. \\
\begin{figure}[h]
    \centering
    \textbf{Q-learning vs. SARSA}\par\medskip
    \includegraphics[width=\columnwidth]{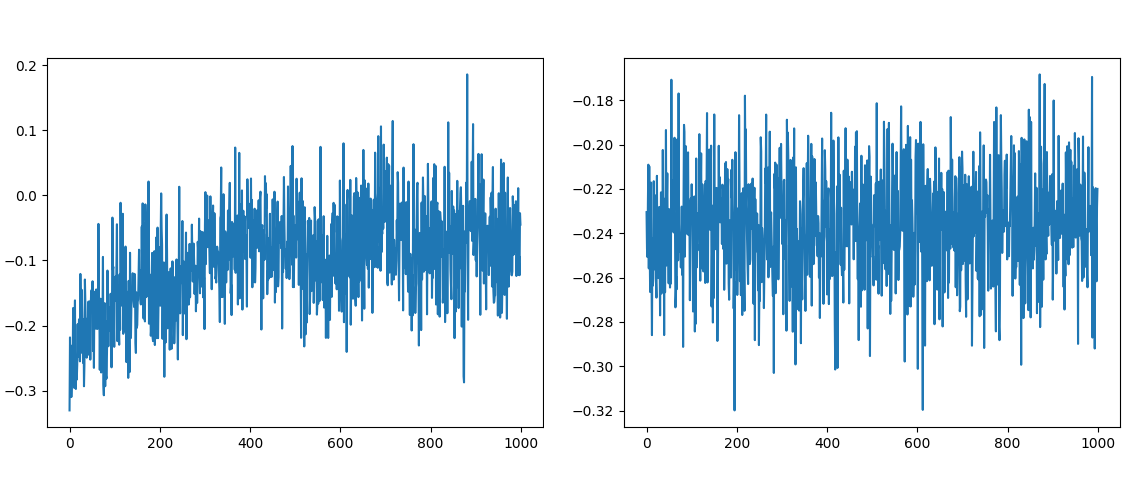}
    \caption{Average rewards for Q-learning (left) and SARSA (right) when run under the same conditional parameters in the HalfCheetah-v2 task from the MuJoCo environment.}
\end{figure}

The following observations listed are from Figure 10:
\begin{itemize}
    \item The plot shows that Q-learning has extremely stochastic behavior, whereas SARSA exhibits more stable behavior over time. This is due to Q-learning's off-policy nature, in which the target and behavior policy are not the same.
    \item The plot shows that with a learning rate of 0.5, both Q-learning and SARSA acquire sub-optimal rewards, while for a learning rate of 1 they perform poorly. This may be linked back to the update equation, in which (1 - learning rate) * Q(s, a) = 0 and we rely only on greediness in Q-learning or randomness in SARSA.  
    \item Q-learning performance improves as the learning rate rises until the learning rate reaches one. However, SARSA's performance is rather stable across all learning rates. 
\end{itemize}

\begin{figure}[h]
    \centering
    \textbf{Q-learning vs. SARSA (Varying learning rate)}\par\medskip
    \includegraphics[width=\columnwidth]{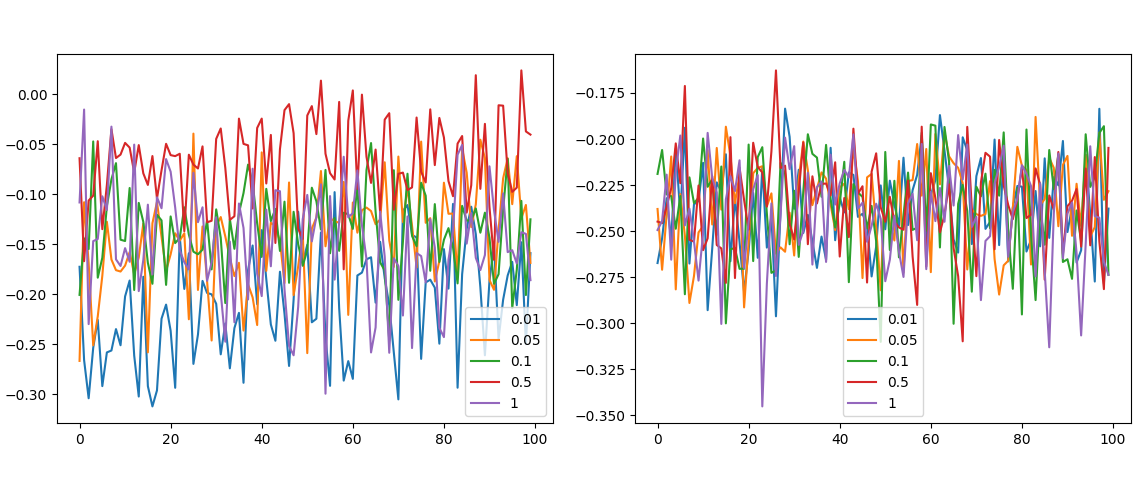}
    \caption{Average rewards for Q-learning (left) and SARSA (right) with $\alpha$ = [0.01, 0.05, 0.1, 0.5, 1] in the HalfCheetah-v2 task from the MuJoCo environment.}
\end{figure}

It was followed by a  Deep Deterministic Policy Gradient (DDPG) network as it was known to perform well on continuous control tasks. The parameter values taken were:
\begin{equation*}
    \gamma = 0.4, \tau = 0.99, \theta = 0.15,
\end{equation*}
\begin{equation*}
    \mu = 0.0, \sigma = 0.3, M' = 10000, n = 100
\end{equation*}
The actor and critic networks were built using two different architectures. The initial architecture included two hidden layers, with the \nth{1} and \nth{2} hidden layers containing 32 and 16 neurons, respectively. The other architecture used 4 hidden layers, each having 32, 64, 32, and 16 neurons for the \nth{1}, \nth{2}, \nth{3}, and \nth{4} hidden layers, respectively. The Adam optimizer was employed for adaptive moment estimation. Mini-batch size = n was used to train the critic network, and mini-batch size = 1 was used to train the actor network.

\subsubsection{Actor network (HalfCheetah-v3 task).}
\begin{center}
    \begin{tabular}{p{3.8cm}p{2.2cm}p{1.5cm}}
        \hline
            \textbf{Layer (type)}   &   \textbf{Output Shape}  &   \textbf{Param \#}  \\
        \hline
        \hline
            dense (Dense)              &  (None, 32)   &    576     \\
            activation (Activation)    &  (None, 32)   &    0       \\
            dropout (Dropout)          &  (None, 32)   &    0       \\
            dense\_1 (Dense)            &  (None, 64)   &    2112   \\
            activation\_1 (Activation)  &  (None, 64)   &    0      \\
            dropout\_1 (Dropout)        &  (None, 64)   &    0      \\  
            dense\_2 (Dense)            &  (None, 32)   &    2080   \\ 
            activation\_2 (Activation)  &  (None, 32)   &    0      \\
            dropout\_2 (Dropout)        &  (None, 32)   &    0      \\
            dense\_3 (Dense)            &  (None, 16)   &    528    \\
            activation\_3 (Activation)  &  (None, 16)   &    0      \\
            dropout\_3 (Dropout)        &  (None, 16)   &    0      \\
            dense\_4 (Dense)            &  (None, 6)    &    102    \\
            activation\_4 (Activation)  &  (None, 6)    &    0      \\
        \hline
        \hline
    \end{tabular}
\end{center}
\subsubsection{Critic network (HalfCheetah-v3 task).}
\begin{center}
    \begin{tabular}{p{3.8cm}p{2.2cm}p{1.5cm}}
        \hline
            \textbf{Layer (type)}   &   \textbf{Output Shape}  &   \textbf{Param \#}  \\
        \hline
        \hline
            input\_2 (InputLayer)        &    [(None, 17)]     &    0       \\
            input\_1 (InputLayer)        &    [(None, 6)]      &    0       \\
            concatenate (Concatenate)    &   (None, 23)        &   0        \\
            dense\_5 (Dense)             &    (None, 32)       &    768     \\
            activation\_5 (Activation)   &    (None, 32)       &    0       \\
            dropout\_4 (Dropout)         &    (None, 32)       &    0       \\
            dense\_6 (Dense)             &    (None, 64)       &    2112    \\
            activation\_6 (Activation)   &    (None, 64)       &    0       \\
            dropout\_5 (Dropout)         &    (None, 64)       &    0       \\
            dense\_7 (Dense)             &    (None, 32)       &    2080    \\
            activation\_7 (Activation)   &    (None, 32)       &    0       \\
            dropout\_6 (Dropout)         &    (None, 32)       &    0       \\
            dense\_8 (Dense)             &    (None, 16)       &    528     \\
            activation\_8 (Activation)   &    (None, 16)       &    0       \\
            dropout\_7 (Dropout)         &    (None, 16)       &    0       \\
            dense\_9 (Dense)             &    (None, 1)        &    17      \\
            activation\_9 (Activation)   &    (None, 1)        &    0       \\
        \hline
        \hline
    \end{tabular}
\end{center}

The number of episodes (epochs) were 10, and each episode had 1000 steps. For the former architecture [32, 16] we observe the average rewards begin with -1200 and after 10 episodes amounted to -300. The latter architecture [32, 64, 32, 16] comparatively performed much better with an initial average reward of -0.621 and converged to -0.401 at the end of 10 episodes.

\begin{figure}[h]
    \centering
    \textbf{DDPG rewards}\par\medskip
    \includegraphics[width=\columnwidth]{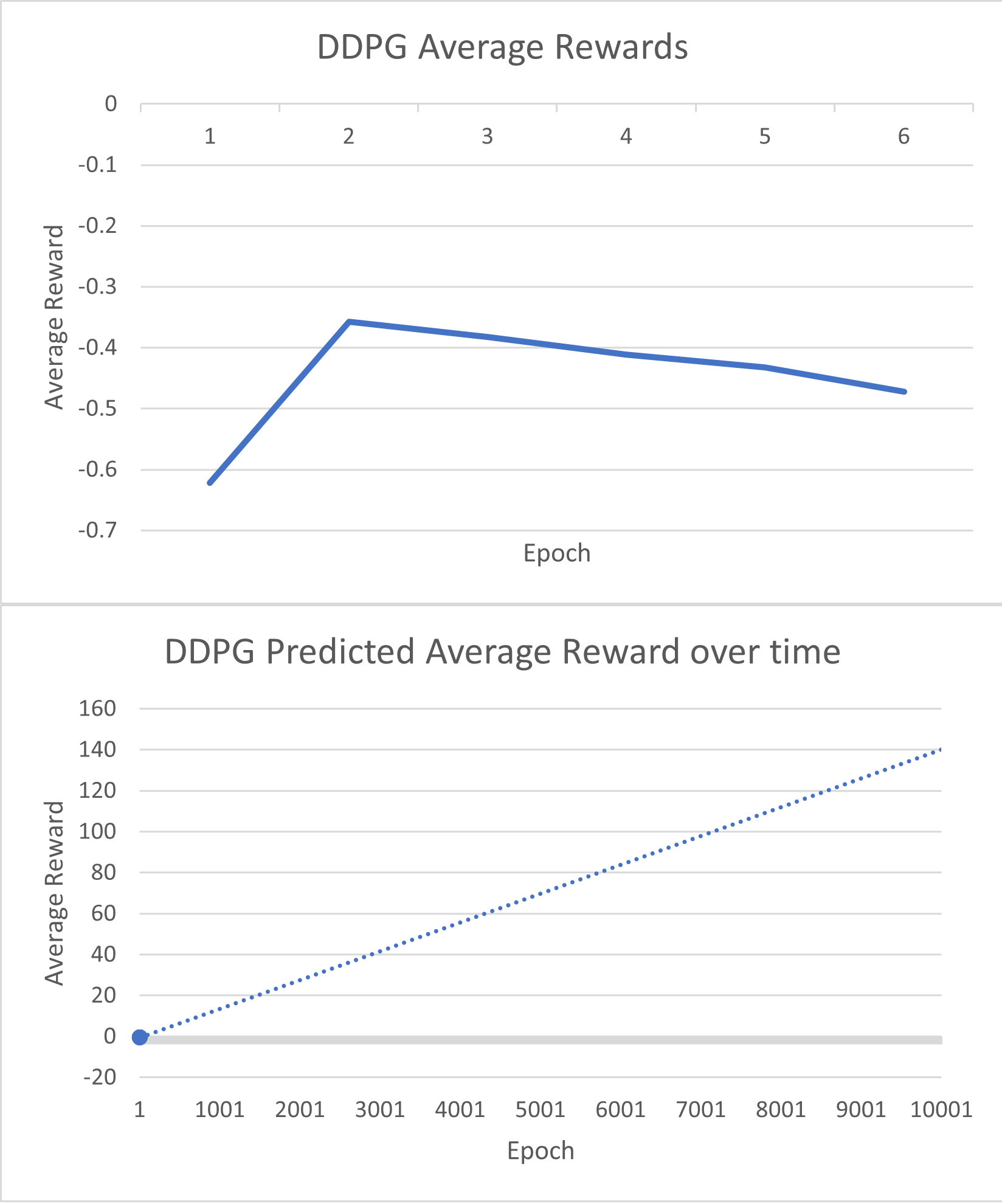}
    \caption{Current average and predicted average rewards for DDPG in HalfCheetah-v2 task with 10000+ iterations.}
\end{figure}

\subsubsection{Improvements.}
The significant improvement involved few changes:
\begin{enumerate}
    \item In contrast to the Linear activation utilized in the actor network of the former architecture, the later used a hyperbolic tangent (tanh) activation function. We assumed that a tanh activation would be meaningful because the values for the actions varied from [-1, 1], and it maps its input from that range.
    \item More hidden layers were added with a 0.2 probability of dropout in the later architecture as compared to fewer layers in the former. As a result, the networks may have learned additional features to better estimate the action-value and policy. The decline in the latter phases of training can be ascribed to overfitting or a greater learning rate leading to overshooting the point of maximum average reward.
\end{enumerate}

\section{Conclusion}
The average rewards received by Q-learning and SARSA for different learning rates are compared in this paper. Q-learning had somewhat better rewards than SARSA, while DDPG, a deterministic policy gradient approach, had even better outcomes than Q-learning and SARSA. There are two things that may be deduced from this.
\begin{enumerate}
    \item Off-policy methods work better compared to on-policy methods on continuous tasks.
    \item Deterministic Policy gradient methods work well in continuous control problems.
\end{enumerate}
By enumerating through the replay buffer, we were able to get minibatches using a non-vectorized version of DDPG. This might be the cause of the algorithm's slowness. In the future, we want to employ vectorized implementations.
Furthermore, given additional simulation time, the anticipated plot in Figure 11 shows that DDPG would eventually lead to greater and better payouts.

\bibliography{ref}
\nocite{*}

\end{document}